# Leveraging Spatial and Semantic Feature Extraction for Skin Cancer Diagnosis with Capsule Networks and Graph Neural Networks

**Kevin Putra Santoso[1]**
5027211030@student.its.ac.id
*Department of Information Technology*
Sepuluh Nopember Institute of Technology

**Raden Venantius Hari Ginardi[2]**
hari@its.ac.id
*Department of Information Technology*
Sepuluh Nopember Institute of Technology

**Rangga Aldo Sastrowardoyo[3]**
5027211059@student.its.ac.id
*Department of Information Technology*
Sepuluh Nopember Institute of Technology

**Fadhl Akmal Madany[4]**
5025221028@student.its.ac.id
*Department of Informatics*
Sepuluh Nopember Institute of Technology

**Abstract**

In the realm of skin lesion image classification, the intricate spatial and semantic features pose significant challenges for conventional Convolutional Neural Network (CNN)-based methodologies. These challenges are compounded by the imbalanced nature of skin lesion datasets, which hampers the ability of models to learn minority class features effectively. Despite augmentation strategies, such as those using Generative Adversarial Networks (GANs), previous attempts have not fully addressed these complexities. This study introduces an innovative approach by integrating Graph Neural Networks (GNNs) with Capsule Networks to enhance classification performance. GNNs, known for their proficiency in handling graph-structured data, offer an advanced mechanism for capturing complex patterns and relationships beyond the capabilities of traditional CNNs. Capsule Networks further contribute by providing superior recognition of spatial hierarchies within images. Our research focuses on evaluating and enhancing the Tiny Pyramid Vision GNN (Tiny Pyramid ViG) architecture by incorporating it with a Capsule Network. This hybrid model was applied to the MNIST:HAM10000 dataset, a comprehensive skin lesion dataset designed for benchmarking classification models. After 75 epochs of training, our model achieved a significant accuracy improvement, reaching **89.23%** and **95.52%**, surpassing established benchmarks such as GoogLeNet (83.94%), InceptionV3 (86.82%), MobileNet V3 (89.87%), EfficientNet-B7 (92.07%), ResNet18 (92.22%), ResNet34 (91.90%), ViT-Base (73.70%), and IRv2-SA (93.47%) on the same dataset. This outcome underscores the potential of our approach in overcoming the inherent challenges of skin lesion classification, contributing to the advancement of image-based diagnosis in dermatology.

## I. BACKGROUNDS

Skin cancer, particularly when left untreated or not managed promptly and effectively, poses a significant health threat, leading to severe complications or increased mortality rates. According to the study by Wilvestra, et. al. [1], some factors such as UV radiation, HPV infection, or genetics may become the cause of this disease. The World Health Organization reports that there are more than two or three million of the non-melanoma and approximately 132.000 melanoma skin cancer cases every year [3]. In Indonesia, the prevalence of skin cancer is the third highest, following cervical cancer and breast cancer [2]. Several common types of this cancer are cell carcinoma, accounting for about 65.5% of all skin cancer types in Indonesia, followed by squamous cell carcinoma at 23%, and malignant melanoma at 7.9%. Roger et al. reported that melanoma skin cancer has a 5-year survival rate of 99% when it has not metastasized and decreases dramatically to 20% at the metastatic stage [5].

Recent advancements in machine learning have enhanced the ability to promptly identify life-threatening cancerous diseases. Certain skin tumors, when discovered early, are non-cancerous and treatable, greatly reducing the risk of malignancy. However, visually inspecting skin cancer in its early stages may pose challenges in accurate analysis due to subjective factors. Hence, there is a necessity for developing more precise and reliable automatic classification techniques to attain a more accurate diagnosis of skin cancer. Recently, deep learning methodologies have notably surpassed traditional machine learning models, particularly in medical image classification, including the classification of skin cancer images.

Several model approaches have been proposed to address this issue. Architectures such as Deep CNN, proposed by Polat et al., have achieved an accuracy of 92.9%, yet they are limited by the fact that these models have not yet proven effective when handling datasets from diverse domains that may exhibit high variance [6]. Rahman, et. al. proposes ensemble deep learning method for skin lesion image classification using the HAM10000 dataset and achieving 88.00% accuracy [7]. Qin et al. suggested a GAN architecture to generate synthetic data, which is then used for training a neural network that has undergone transfer learning [8]. This model achieved a significant accuracy of 95.2%, but it has the drawback that the content of the synthetic dataset produced by the GAN is not complex and varied enough compared to the original data [4]. Another approach by Gu, et. al. (2020) tries to improve model generalization ability and robustness [10]. In his research, Gu proposed transfer learning and adversarial learning in skin disease classification to improve the generalization ability of models to new samples and reduce cross-domain shift [10]. The model used in this approach also gives significant accuracy of 90.9%. However, this model accuracy suffers when the data domain and target domain are significantly different [4][10]. These research examples demonstrate that the challenge of skin cancer image classification lies in the complexity of the features within its data, especially on its semantic and spatial factors that is unique with one another.

Research on Graph Neural Networks on image classification has rapidly emerged in the past few years. Vasudevan, et. al. introduced WaveMesh, a wavelet-based superpixeling algorithm, which generates image-specific, multiscale superpixels [11]. These superpixels are structurally distinct from traditional methods, providing a more content-driven graph representation of images. The research leverages SplineCNN, a cutting-edge GNN for image graph classification, to evaluate the performance of WaveMesh against conventional superpixel methods. The results indicated that WaveMesh, especially with the WavePool pooling strategy, achieved competitive performance of 98.68% accuracy on MNIST digit dataset [11]. Similarly, Chatzianastasis et al. utilized an Explainable Multilayer Graph Neural Network (EMGNN) for identifying cancer genes using gene–gene interaction networks and multi-omics data [12]. EMGNN consistently outperformed existing methods, showing an average 7.15% improvement in area under the precision–recall curve, while providing valuable biological insights and explanations for its predictions. Ramirez et al. focused on using graph convolutional neural networks (GCNN) for classifying tumor and non-tumor samples into 33 cancer types or as normal based on gene expression data [13]. Their GCNN models achieved high prediction accuracies (89.9–94.7%) across 34 classes, and in silico gene-perturbation experiments identified 428 marker genes driving cancer classification. This research demonstrated the potential of GNNs in accurate cancer classification and genomic mechanism understanding. These advancements collectively highlight the potential of GNNs not only in accurate cancer classification but also in enhancing image-based diagnosis, making them promising for applications like skin cancer diagnosis.

Capsule Network is a type of neural network that is suitable in handling this challenge. Given the limitations of traditional deep learning approaches, the Capsule Network emerges as a promising solution for the challenges in skin cancer image classification. This novel architecture, first introduced by Sabour, Hinton, et al. in 2017, is designed to capture the hierarchical relationships and spatial hierarchies between the parts of an object, making it particularly effective for tasks requiring a nuanced understanding of an image's spatial and semantic features [14]. Capsule Networks differ from conventional Convolutional Neural Networks (CNNs) by processing data in "capsules", small groups of neurons that specialize in recognizing various aspects of an image, such as texture, shape, or orientation. These capsules are capable of capturing the spatial relationships between features, allowing the network to maintain a high degree of accuracy even when an image is viewed from different angles or in various lighting conditions [19]. This property is especially beneficial for skin cancer diagnosis, where the appearance of lesions can vary significantly depending on the angle of the image capture and the lesion's location on the body.

In response to the demonstrated potential of Graph Neural Networks (GNNs) for their enhanced flexibility and generalization capabilities in image processing, this study seeks to apply GNNs in the classification of skin cancer. Specifically, we employed the Vision Graph Neural Network (ViG) framework as proposed by Kai et al., leveraging its proven efficacy in image classification as evidenced by its performance on the

ImageNet dataset [15]. This research further explores the integration of ViG with Capsule Networks to evaluate the potential impact on the performance of GNNs. By replacing the traditional final prediction layer, which typically consists of multiple convolutional and pooling layers along with batch normalization, with a Capsule Network layer, we aim to address the limitations identified by Hinton et al. regarding the pooling layers' inadequacy in preserving spatial relationships between features [14]. ViG is available in two configurations: isotropic and pyramid. Each configuration comes in four sizes: tiny, small, medium, and large. According to Kai, et. al.'s research, the pyramid configuration consistently outperforms its isotropic counterpart across all sizes [15]. Given the constraints of our resources, our study focused on the 'tiny' variant of the pyramid model, which was selected based on its documented superior performance on the ImageNet dataset relative to the isotropic version.

## II. RELATED WORKS

The study by Alam et al., conducted in 2022, developed an efficient deep learning-based classifier for skin cancer utilizing the RegNetY-320 model [16]. To address the imbalance across various skin cancer classes, the team employed data augmentation techniques. This approach allowed the model to achieve a peak accuracy of 91% and an average F1-score of 88.1% on the HAM10000 dataset. Another research with the same dataset, Datta, et. al. in 2021 proposes Soft-Attention Mechanism approach in the IRv2-SA model; combining the Inception Network and ResNet [17]. The soft-attention module enables IRv2-SA to focus more on important features on the skin lesion images using soft-attention maps and dispose irrelevant feature such as hair and veins. This model gives 93.47% accuracy [17]. Zhao, et. al. in 2021 developed a classification system utilizing a skin lesion augmentation technique through Style-based GAN (SLA-StyleGAN) [18]. This approach reached an accuracy rate of 93.64% when tested on both the ISIC2018 and ISIC2019 datasets. The research of skin cancer classification extended to the adoption of the famous Vision Transformer (ViT) architecture through Pedro, et. al. research [20]. Pedro, et. al. investigates the role of attention and self-attention within neural networks, particularly for classifying skin lesions using the Skin Cancer MNIST:HAM10000 dataset. The study examines two approaches: attention modules, which adjust feature weighting within layers of the ResNet architecture, and self-attention mechanisms, initially developed for Natural Language Processing but increasingly applied in Computer Vision. The findings indicate that while attention modules can enhance convolutional neural network (CNN) performance variably, self-attention mechanisms consistently improve outcomes, even with fewer parameters. As of the quantitative results, this approach achieved 62.20% accuracy adopting the vanilla attention mechanisms on ResNet and 73.70% accuracy adopting the self-attention mechanisms on ViT [20].

All the studies mentioned exhibit a common characteristic: the application of deep, stacked convolutional layers. The reduced accuracy in some architectures can be attributed to issues such as vanishing gradients or the models' insufficient capability to capture complex features. The research we propose adopts an alternative strategy, utilizing a Graph Neural Network (GNN)-based model that is enhanced with a Capsule Network in its terminal layer. This design aims to mitigate the loss of spatial feature information, thereby significantly improving the model's ability to thoroughly learn features from each image class.

## III. MODEL ARCHITECTURE

Graph Neural Networks (GNNs) were initially introduced by Gori, et. al. in 2005 [21][22]. GNNs represent a deep learning model where the configuration of artificial neurons is structured as a graph []. A graph is defined as $\mathcal{G} = (\mathcal{V}, \mathcal{E})$, where $\mathcal{V}$ represents a set of vertex embeddings $\vec{v} \in \mathcal{V}$ and $\mathcal{E}$ denotes a set of edges $\vec{e}_{ij} \in \mathcal{E}$ connecting vertices $\vec{v}_i, \vec{v}_j \in \mathcal{V}$. Each vertex that maintains a relationship with a specific vertex is referred to as a neighboring vertex. The set that contains all neighboring vertices of $\vec{v}_i \in \mathcal{V}$ is denoted as $\mathcal{N}(\vec{v}_i)$. GNNs operate by updating the parameters of each existing vertex through the use of an aggregation operator. In general, this operation is mathematically defined as follows.

$$\mathcal{G}_{\ell+1} = \mathcal{F}(\mathcal{G}_\ell, W_\ell) \qquad (3.1)$$

where $\mathcal{F}: \mathcal{G} \to \mathcal{G}$ denotes the graph update function operation. In this process, each vertex is updated using the formula:

$$\vec{v}_i := \phi(v_i, \rho(v_i, \mathcal{N}(v_i), W_\rho), W_\phi). \qquad (3.2)$$

Here, $\phi$ is a non-linear function and $\rho$ is an aggregation function involving vertex $\vec{u}$, which is a neighbor of $\vec{v}_i$. The non-linear function can be ReLU, Hardswish, Tanh, or an Artificial Neural Networks (ANNs) such as a Multilayer Perceptron (MLP), among the other non-linear functions. On the other hand, the aggregation function may take the form of an average, ANN (MLP), sequential neural networks (such as RNN, LSTM, etc.), maximum value, and so forth [23].

Initially, Graph Neural Networks (GNNs) were commonly used for node-level classification or recommendation systems although it has limitations in its implementation on images [23]. This limitation stemmed from the complex structure of images—consider a $256 \times 256$ image, representing it as a graph would entail having as many as 65,536 vertices. As the size of the image increases, the complexity of the graph escalates with the product of the image's width and height, implying a complexity of $O(HW)$. Despite these challenges, there are advantages to applying graph representations to images. Firstly, graphs are a ubiquitous data structure, allowing grids or sequences to be viewed as specialized forms of graphs [15]. Secondly, graphs offer greater flexibility than grids or sequences in modelling complex objects, as is often required in images. Thirdly, an object can be decomposed into compositions of other objects (for example, a human body can be broken down into limbs), and graph structures can be utilized to construct relationships among these parts. Fourth and most importantly, current research on GNNs is focused on these tasks [15].

To address the complexity issues associated with images, Vision GNN (ViG) employ a strategy where patches from the image are extracted and then transformed into vector feature representations. For an image of dimensions $H \times W \times 3$, it is divided into $N$ patches by applying a transformation that converts each patch into a feature vector $X = [\vec{x}_1, \ldots, \vec{x}_N]$ where $D$ is the feature dimension and $N$ represents the number of vertices in the graph representation. The process proceeds through three levels: feature map extraction, graph-level processing, and spatial feature learning through a capsule network as shown in the Figure 3.1 below.

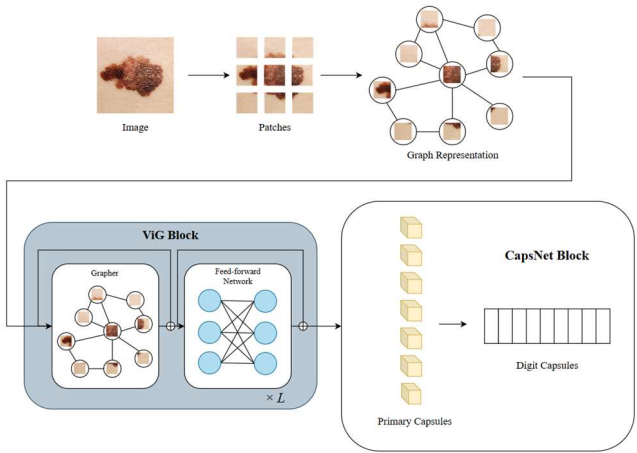

**Figure 3.1** The ViG-CapsNet Architecture

Each component will be explained in detail in the following sections.

### 3.1 Feature Map Extraction (Stem)

An image of size $H \times W \times 3$ is processed as a tensor $X \in \mathbb{R}^{3 \times H \times W}$. Features from the tensor $X$ are extracted through a convolution layer, resulting in a feature map $F' \in \mathbb{R}^{C \times H \times W}$. In general, this operation is defined mathematically as follows.

$$F' = f_{\text{Conv}}(X) = \sigma(WX), \quad W \in \mathbb{R}^{C \times 3} \quad (3.3)$$

where $f_{\text{Conv}} : \mathbb{R}^{3 \times H \times W} \to \mathbb{R}^{C \times H \times W}$ is the function denoting the transformation by convolution in general, and $\sigma$ denotes the activation function. This operation occurs in the first part of the ViG architecture, referred to as the Stem.

### 3.2 Graph-level Processing (Backbone)

The feature map $F'$ obtained from extraction in the Stem is passed to the Backbone section. In this part, feature transformations and graph-level processing occur. In general, to obtain the graph representation, the feature map $F'$ is divided into $N$ patches by applying a transformation that converts each patch into a feature vector $\vec{x}_i \in \mathbb{R}^D$ and contain it in a matrix $X = [\vec{x}_1, \ldots, \vec{x}_N]^T$ where $D$ is the feature dimension and $N$ represents the number of vertices in the graph representation. These features can be viewed as a set of unordered vertices denoted as $\mathcal{V} = \{v_1, \ldots, v_N\}$. Subsequently, the K nearest neighbors are computed for each vertex, resulting in $\mathcal{N}(v_i)$ as well as each directed edge $e_{ji}$ from $v_j$ to $v_i$ for all $v_j \in \mathcal{N}(v_i)$. From this process, the graph representation $\mathcal{G} = (\mathcal{V}, \mathcal{E})$ is obtained.

Practically, the feature vector $X \in \mathbb{R}^{N \times D}$ is then processed by a graph convolutional layer that can exchange information between vertices by aggregating features from its neighbouring vertices. In this case, the update and aggregation functions are defined as follows.

$$\vec{x}'_i := \phi(\vec{x}_i, \rho(\vec{x}_i, \mathcal{N}(\vec{x}_i), W_\rho), W_\phi) \quad (3.4)$$

where $\mathcal{N}(\vec{x}_i)$ denotes the set of neighbouring vertices of $\vec{x}_i$. Subsequently, the operation of max-relative graph convolution is applied, in which the nonlinear function and its aggregation implement:

$$\begin{aligned} \rho(\cdot) = \vec{x}''_i = [\vec{x}_i, \max(\{\vec{x}_j - \vec{x}_i | j \in \mathcal{N}(\vec{x}_i)\})] \\ \phi(\cdot) = \vec{x}'_i = \vec{x}''_i W_\phi \end{aligned} \quad (3.5)$$

where $W_\phi \in \mathbb{R}^{D \times D}$ is a weight matrix of the update function. The aggregated vector $\vec{x}''$ is then divided into $h$ parts, referred to as heads. Each head is updated with a different weight matrix. After each head is updated in parallel with its weight matrix, the outputs are concatenated back together, resulting in the final value

$$\vec{x}'_i = \left[\vec{h}_1 W_\phi^{(1)}, \vec{h}_2 W_\phi^{(2)}, \ldots, \vec{h}_h W_\phi^{(h)}\right] \quad (3.6)$$

In summary, the transformation of $X = [\vec{x}_1, \ldots, \vec{x}_N]$ to be $X' = [\vec{x}'_1, \ldots, \vec{x}'_N]$ is denoted as $G(X)$.

The $G$ operation is used to construct a single ViG block, where each block consists of two components: the Grapher and the Feed-forward Network (FFN). The Grapher module is expressed as:

$$Y = \sigma(G(XW_{in}))W_{out} + X \quad (3.7)$$

where $Y \in \mathbb{R}^{N \times D}$, $W_{in}$ and $W_{out}$ are the weight matrices of this layer, and $\sigma$ is the activation function. The output from the Grapher module is then passed to the FFN module, which is defined as:

$$Z = \sigma(YW_1)W_2 + Y \quad (3.8)$$

where $Z \in \mathbb{R}^{N \times D}$, $W_1$ and $W_2$ are the weight matrices of this layer. The purpose of implementing the FFN module is to prevent the condition of oversmoothing [24], which is a situation where important information from the features of each vertex disappears due to excessive aggregation operations. The output from a single ViG block will have dimensions $\mathbb{R}^{C' \times H' \times W'}$, where $C', H',$ and $W'$ are the number of output channels and the size of the output image's width and height.

### 3.3 Spatial Feature Learning Through the Capsule Network (Predictor)

Normally, the predictor layer is the last layer of the ViG, where this layer consists of a number of convolutional layers with activation functions similar to the FFN [15]. However, in this part, we apply a Capsule Network (CapsNet) to enhance the model's performance in learning spatial features [14][19]. The CapsNet layer returns 7-classes vector $\vec{v} \in \mathbb{R}^d$ where $d$ denotes the final capsule size. It is defined as

$$V = \{\vec{v_1}, \ldots, \vec{v_c}\} \quad (3.9)$$

where $c$ denotes the number of classes. We obtain the class by taking the argument of the maximum value of Euclidean norm for each vector. Mathematically, this is defined as

$$\|v_i\| = \sqrt{v_{i1}^2 + v_{i2}^2 + \cdots + v_{id}^2}, i \in \{0, \ldots, c\} \quad (3.10)$$

$$\hat{c} = \underset{V}{\mathrm{argmax}}\{\|\vec{v_1}\|, \ldots, \|\vec{v_c}\|\} \quad (3.11)$$

where $\hat{c}$ is the output class. The loss of this network is calculated using the margin loss function. This loss function helps the capsule network learn to produce high-magnitude outputs for the correct class capsules and low-magnitude outputs for the incorrect class capsules. This enables the network to make confident predictions and achieve good classification performance. This function calculated by the formula

$$L_k = T_k \max(0, m^+ - \|\mathbf{v_k}\|)^2 \\ + \lambda(1 - T_k)\max(0, \|\mathbf{v_k}\| - m^-)^2 \quad (3.12)$$

where the $L_k$ denotes as single capsule k, $T_k$ is the target label for capsule k, the range of it values is set for 1 to be active for the correct class, and 0 for otherwise, $\|\mathbf{v_k}\|$ is the magnitude of the output vector of capsule k that indicates how strongly the capsule believes the input image belongs to the class it is responsible for, and there is two margin values $m^+$ which is the margin for the correct class that typically set to 0.9, and $m^-$ which is the margin for the incorrect class, which is usually set at 0.1, as defined in the original CapsNet paper [14]. The use of $m^+$ margin values ensures that the capsule output vector length for the correct class exceeds 0.9, while $m^-$ ensures that the vector lengths for incorrect classes stay below 0.1, effectively pushing the network to make clear distinctions between classes. The $\lambda$ is a down weighting factor for the loss of incorrect classes (typically set to 0.5). The first term, $T_k \max(0, m^+ - \|\mathbf{v_k}\|)^2$ penalizes the network when the magnitude of the output vector $\|\mathbf{v_k}\|$ for the correct class capsule is less than $m^+$. This encourages the network to learn capsule outputs with high magnitudes for the correct class. The second term, $\lambda(1 - T_k)\max(0, \|\mathbf{v_k}\| - m^-)^2$, penalizes the network when the magnitude of the output vector for an incorrect class capsule is greater than $m^-$. This encourages the network to learn capsule outputs with low magnitudes for the incorrect classes. During training, the capsule network aim to minimize the total margin loss by adjusting its weights. This process encourages the capsule network to produce outputs with a high magnitude (greater than $m^+$ for the correct class and a low magnitude (less than $m^-$) for the incorrect classes. The lower the total margin loss, the better the performance and generalization of the capsule network.

### 3.4 Architecture Design

The Table 3.1 below describes the detailed settings of the models used in this experiment. $D$ stands for feature dimension, $E$ stands for ratio rate in $FFN$, $K$ stands for the number of neighbors in the Graph Convolutional Network, $H \times W$ denotes the input size, PViG stands for Pyramid Vision GNN (ViG).

**Table 3.1** Detailed settings of our models.

| Stage | Output size | PViG-Tiny | PViG-CapsNet-Tiny |
|---|---|---|---|
| Stem | $\frac{H}{4} \times \frac{H}{4}$ | Conv2D $\times$ 3 | Conv2D $\times$ 3 |
| Stage 1 | $\frac{H}{4} \times \frac{H}{4}$ | $\begin{bmatrix} D=48 \\ E=4 \\ K=9 \end{bmatrix} \times 2$ | $\begin{bmatrix} D=48 \\ E=4 \\ K=9 \end{bmatrix} \times 2$ |
| Downsample | $\frac{H}{8} \times \frac{H}{8}$ | Conv2D | Conv2D |
| Stage 2 | $\frac{H}{8} \times \frac{H}{8}$ | $\begin{bmatrix} D=96 \\ E=4 \\ K=9 \end{bmatrix} \times 2$ | $\begin{bmatrix} D=96 \\ E=4 \\ K=9 \end{bmatrix} \times 2$ |
| Downsample | $\frac{H}{16} \times \frac{H}{16}$ | Conv2D | Conv2D |
| Stage 3 | $\frac{H}{16} \times \frac{H}{16}$ | $\begin{bmatrix} D=240 \\ E=4 \\ K=9 \end{bmatrix} \times 2$ | $\begin{bmatrix} D=240 \\ E=4 \\ K=9 \end{bmatrix} \times 2$ |
| Downsample | $\frac{H}{32} \times \frac{H}{32}$ | Conv2D | Conv2D |
| Stage 4 | $\frac{H}{32} \times \frac{H}{32}$ | $\begin{bmatrix} D=384 \\ E=4 \\ K=9 \end{bmatrix} \times 2$ | $\begin{bmatrix} D=384 \\ E=4 \\ K=9 \end{bmatrix} \times 2$ |
| Head | $1 \times 1$ (Original), $d \times 1$ (CapsNet) | Pooling & MLP | Capsule Network |
| Parameters (M) | | 9.54 | 9.48 |
| FLOPs (B) | | 2.07 | 4.17 |

Based on this table, we can see clearly that PViG-CapsNet-Tiny contains less parameters but higher floating point operations (FLOPs) than PViG-Tiny.

## IV. EXPERIMENTS AND RESULTS

### 4.1 Experimental Settings

**Datasets.** We implemented the Pyramid-ViG-Tiny and Pyramid-ViG-CapsNet-Tiny models using PyTorch 2.2 and trained it on a single server with Intel I7 CPU and NVIDIA RTX3070 GPU. The training takes 75 epochs and 8 hours to complete.

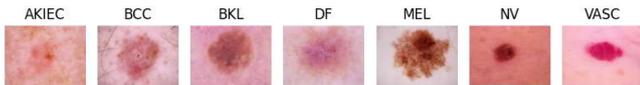

**Figure 4.1** Sample image for each skin lesion type.

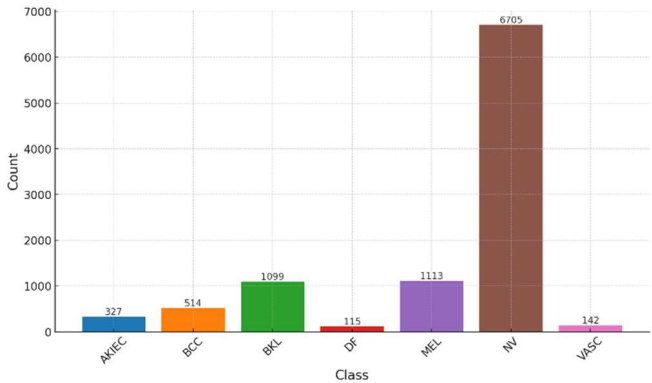

**Figure 4.2** Distribution of skin lesion classes in HAM10000 dataset

The MNIST:HAM10000 dataset is used during in this study [25]. This dataset consisted of 10015 dermatoscopic images in PNG format with a size of $450 \times 600$. There are seven types of skin lesions: actinic keratosis/intraepithelial carcinoma (AKIEC), basal cell carcinoma (BCC), melanoma (MEL), melanocytic nevi (NV), and vascular lesions (VASC). One challenge of this dataset is that it is highly imbalanced, as illustrated in the Figure 4.2. Based on this figure, NV holds about 66.9% of the dataset. In the other hand, the DF class is the minority class which only holds about 1.14% of the whole dataset.

Every image is resized to $256 \times 256$ and augmented with random crops and flips. The evaluation metrics used in this study are Accuracy and F1-Score.

$$Accuracy = \frac{TP+TN}{TP+TN+FP+FN} \quad (4.1)$$

$$F1\ Score = \frac{2 \cdot Precision \cdot Recall}{Precision + Recall} \quad (4.2)$$

**Baselines.** We have opted to use the Tiny Pyramid model due to our resource limitations and since Pyramid models performs better than the isotropic models as mentioned in the original ViG paper [15]. The isotropic architecture maintains the feature size through the network's computational core which allows for easy scaling and hardware acceleration. On the other hand, pyramid neural networks gradually reduce the spatial size of feature maps as the network deepens. This leverages the scale-invariant property of images to produce multi-scale features. In this work, we compare our PViG-CapsNet-Tiny model with the original PViG-Tiny and several other models used in the past research such as IRv2-SA, GoogLeNet, InceptionV3, MobileNet, EfficientNet-B7, ViT-Base, and ResNet [17][20][26][27][28][29].

**Hyperparameters.** For all ViG models, we implemented dilated aggregation in the Grapher module with a dilated rate of $\ell/4$ for the $\ell$-th layer. GELU was used as the activation function in equations 3.7 and 3.8. Details regarding other hyperparameters are presented in Table 4.1 below.

**Table 4.1** Hyperparameters used for training PViG-Tiny and PViG-CapsNet-Tiny

| Hyperparameters | PViG-Tiny | PViG-CapsNet-Tiny |
|---|---|---|
| Epochs | | 75 |
| Optimizer | | AdamW [30] |
| Batch size | | 32 |
| Start learning rate | | $1 \times 10^{-6}$ |
| Learning rate scheduler | | Cosine |
| Learning rate | | $2 \times 10^{-3}$ |
| Warmup epochs | | 20 |
| $m^+$ | | 0.9 |
| $m^-$ | None | 0.1 |
| $\lambda$ | | 0.5 |

It should be noted that the hyperparameters $m^+, m^-, \lambda$ are parameters for the margin loss function implemented in PViG-CapsNet-Tiny. For the PViG-Tiny model, the loss function used is Cross Entropy without additional hyperparameters [31].

### 4.2 Skin Lesion Type Classification Result

During the training process, PViG-CapsNet-Tiny demonstrated superior learning capabilities compared to PViG-Tiny, as observed from the loss graphs.

According to Figure 4.2, it is evident that PViG-CapsNet-Tiny achieved significantly lower minimum loss values than PViG-Tiny for both training and validation phases.

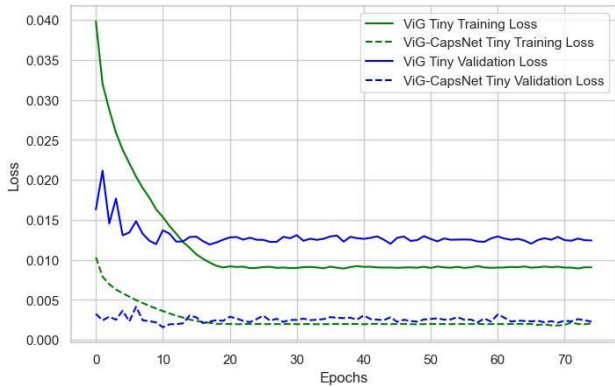

**Figure 4.2** Comparation of training and validation losses of each model over training epochs.

Additionally, PViG-CapsNet-Tiny also managed to achieve a relatively higher validation set accuracy compared to the training set accuracy during its training phase, in contrast to PViG-Tiny, which reached the same equilibrium after the 20th epoch and stagnated at an accuracy rate of around 89%. This demonstrates that the utilization of CapsNet enhances the learning capabilities of PViG. Figure 4.3 illustrates the comparative progression of training set and validation set accuracy across epochs.

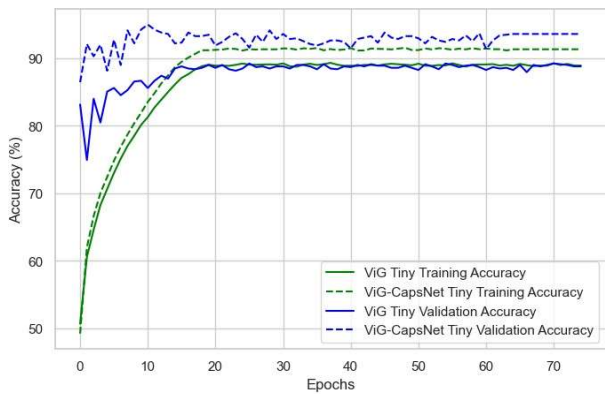

**Figure 4.3** Comparation of training and validation accuracy of each model over training epochs

Table 4.2 provides complete information regarding the performance comparison between these models.

**Table 4.2** PViG-Tiny and PViG-CapsNet-Tiny comparison among several state-of-the-arts.

| Model | ACC (%) | Params (M) | FLOPs (G) |
|---|---|---|---|
| IRv2-SA | 93.47 | 47.5 | 25.46 |
| GoogLeNet | 83.94 | 5.98 | 1.58 |
| InceptionV3 | 86.82 | 22.8 | 5.73 |
| MobileNet | 89.87 | 1.53 | 0.12 |
| EfficientNet-B7 | 92.07 | 66.3 | 37.75 |
| ResNet18 | 92.22 | 91.42 | 4.09 |
| ResNet34 | 91.90 | 95.43 | 4.09 |
| ViT-Base | 73.70 | 85.80 | 66.4 |
| PViG-Tiny | 89.23 | 9.54 | 2.07 |
| **PViG-CapsNet-Tiny** | **95.52** | 9.48 | 4.17 |

The results of the testing process demonstrate that the performance of PViG-Tiny gives 89.23% accuracy, yet elevated significantly to 95.52% when CapsNet (PViG-CapsNet-Tiny) is employed. It can bee seen that PViG-CapsNet-Tiny outperforms several mentioned state-of-the-art models.

**Table 3.3** Evaluation metrics of PViG-CapsNet-Tiny, PViG-Tiny, and IRv2-SA for each skin lesion type on the test set

| Class | Recall | | | F1-Score | | |
|---|---|---|---|---|---|---|
| | PViG-CapsNet-Tiny | PViG-Tiny | IRv2-SA | PViG-CapsNet-Tiny | PViG-Tiny | IRv2-SA |
| AKIEC | **0.692** | 0.462 | 0.520 | **0.750** | 0.521 | 0.690 |
| BCC | 0.867 | 0.833 | **0.880** | 0.788 | 0.793 | **0.880** |
| BKL | 0.800 | 0.546 | **0.830** | **0.876** | 0.607 | 0.770 |
| DF | **0.667** | **0.667** | 0.170 | **0.800** | 0.571 | 0.290 |
| MEL | **0.692** | 0.436 | 0.650 | **0.771** | 0.465 | 0.660 |
| NV | **1.000** | 0.968 | 0.980 | **0.986** | 0.957 | 0.980 |
| VASC | 0.909 | **1.000** | **1.000** | 0.952 | 0.846 | **1.000** |

Furthermore, we compared PViG-CapsNet-Tiny, PViG-Tiny, and IRv2-SA which having the second highest accuracy based on the mentioned evaluation metrics. Table 3.3 above provides the comparison of each model.

## V. DISCUSSIONS

Based on the result shown, it has been proven that Pyramid ViG (PViG) is quite effective in classifying types of skin lesions. However, its effectiveness can be significantly enhanced by the use of a capsule network after features are processed by the Backbone component, where graph-level feature processing takes place. Before the implementation of the capsule network, PViG could only achieve a maximum accuracy of 89.23%. This result still did not surpass some of the existing models. However, the application of the capsule network increased the accuracy to 95.52%, enabling it to exceed all the benchmark models that were tested.

Furthermore, it can also be observed that the validation accuracy of PViG-CapsNet-Tiny achieves a superior figure compared to PViG-Tiny in a shorter period and converges at a higher value. Table 5.1 provides details on the best top-$k$ performance during $k$ epochs in the validation set, based on the data presented in Figure 4.3.

**Table 5.1** The top-$k$ validation accuracy of $k$ epochs for PViG-Tiny and PViG-CapsNet-Tiny

| | PViG-Tiny | PViG-CapsNet-Tiny |
|---|---|---|
| **Top-1** | 83.15 | 86.46 |
| **Top-5** | 85.07 | 92.11 |
| **Top-10** | 86.67 | 94.48 |
| **Top-20** | 89.02 | **95.52** |
| **Top-50** | **89.23** | 95.52 |
| **Top-75** | 89.23 | 95.52 |

The accuracy of PViG-CapsNet-Tiny reaches its peak before the 20th epoch, whereas PViG-Tiny requires more than 20 epochs to achieve its highest accuracy.

This adds evidence that CapsNet is highly effective in significantly enhancing the learning capability of PViG-Tiny.

Based on the existing state-of-the-art comparisons, IRv2-SA occupies the second position in terms of highest accuracy among other models. We selected this model as the primary comparator against the two models we developed and tested, namely PViG-Tiny and PViG-CapsNet-Tiny, in aspects of recall and F1-score. According to the data listed in Table 3.3, it can be seen that the recall of PViG-CapsNet-Tiny is superior to both PViG-Tiny and IRv2-SA in several classes, such as AKIEC, DF, MEL, and NV. One interesting finding is the superiority of PViG-CapsNet-Tiny in detecting the DF class, which has the least amount of data compared to other classes. This advantage is also supported by a relatively higher F1-Score compared to other models. This indicates the ability of PViG-CapsNet-Tiny to learn well from minority class data, which aligns closely with clinical objectives in medical practice. By using the F1-Score as an indicator, it demonstrates the model's ability to sensitively detect all possible cases, thereby minimizing the possibility of missing critical cases. However, there is room for improvement for this model to be more viable in a medical context. One approach that can be taken is by implementing larger versions of the model, such as PViG-CapsNet-Small, PViG-CapsNet-Medium, or PViG-CapsNet-Large, to further increase its accuracy and reliability.

## VI. CONCLUSIONS

In conclusion, this research demonstrates that Pyramid ViG (PViG), when augmented with a capsule network, significantly improves its effectiveness in classifying various types of skin lesions using the imbalanced MNIST:HAM10000 dataset. The implementation of the capsule network not only enhances PViG's accuracy from 89.23% to 95.52%, surpassing existing models, but also shows superior performance in terms of validation accuracy and faster convergence when compared to its smaller counterpart, PViG-Tiny. The enhanced model, PViG-CapsNet-Tiny, notably excels in detecting minority classes such as DF, indicating its potential for clinical application by effectively learning from limited data and ensuring sensitive detection of all possible cases, thus minimizing the risk of overlooking critical conditions.

However, despite these promising results, there exists potential for further enhancements to make the model more suited for medical contexts. Exploring larger variations of the model, such as PViG-CapsNet-Small, Medium, or Large, could potentially lead to improvements in accuracy and reliability, which we are going to explore in the future. We will also continue this research to discover the ability of the ViG-CapsNet and PViG-CapsNet model in generalization in the future.